\documentclass[10pt,twocolumn,letterpaper]{article}

\usepackage{wacv}
\usepackage{times}
\usepackage{epsfig}
\usepackage{graphicx}
\usepackage{amsmath}
\usepackage{amssymb}
\usepackage{algorithm}
\usepackage{algorithmic}
\usepackage[accsupp]{axessibility}
\usepackage{booktabs} 
\usepackage{amsmath}
\usepackage{mathabx}

\def\wacvPaperID{19} 

\wacvfinalcopy 

\usepackage[pagebackref=true,breaklinks=true,colorlinks,bookmarks=false]{hyperref}

\begin{document}


\title{Utilizing Network Features to Detect Erroneous Inputs}




\author{Matt Gorbett\\
Colorado State University\\
{\tt\small matt.gorbett@colostate.edu}
\and
Nathaniel Blanchard\\
Colorado State University\\
{\tt\small nathaniel.blanchard@colostate.edu}
}

\maketitle
\thispagestyle{empty}

\begin{abstract}
Neural networks are vulnerable to a wide range of erroneous inputs such as corrupted, out-of-distribution, misclassified, and adversarial examples. Previously, separate solutions have been proposed for each of these faulty data types, however, in this work we show that a collective set of inputs with variegated data quality issues can be jointly identified with a single model.  
Specifically, we train a linear SVM classifier to detect four types of erroneous data using the hidden and softmax feature vectors of pre-trained neural networks. 
Our results indicate that these faulty data types generally exhibit linearly separable activation properties from correctly processed examples.  
We are able to identify erroneous inputs with an AUROC of 0.973 on CIFAR10, 0.957 on Tiny ImageNet, and 0.941 on ImageNet.  
We experimentally validate our findings across a diverse range of datasets, domains, and pre-trained models. 
\end{abstract}

\section{Introduction}

Humans are capable of adapting to diverse types of data in ways machine learning models cannot~\cite{geirhos_generalisation_2018}.
While the human visual system is able to generalize across varying image representations, such as different Instagram filters, deep learning classifiers misbehave when presented with image corruptions \cite{dodge_study_2017,hendrycks_benchmarking_2019}, adversarial examples \cite{szegedy_intriguing_2014}, and previously unseen classes \cite{bendale_towards_2015, hendrycks_deep_2019, scheirer_2013}. 

\begin{figure}[!ht]\centering
 \label{teaser}
 \includegraphics[width=1\linewidth]{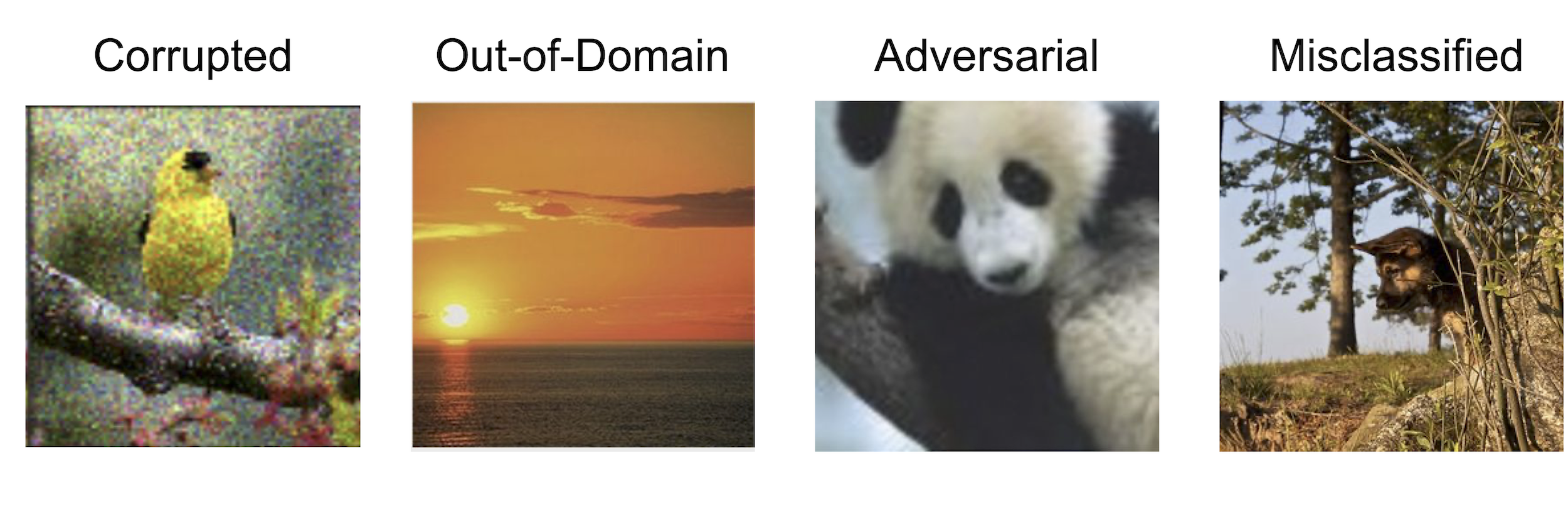}

 \caption{Modern neural network models have state-of-the-art performance on image tasks like object classification, however, these networks are vulnerable to erroneous inputs i.e., images that are corrupted, out-of-domain, adversarial, or misclassified. Previous research has focused on separate solutions to defending against each type of erroneous input, but we show that all erroneous inputs can be jointly identified with a linear SVM trained on a given network's activations. 
}
\end{figure}

Figure \ref{teaser} shows an example of each type of erroneous input: image corruptions \cite{dodge_study_2017,hendrycks_benchmarking_2019},  previously unseen classes \cite{bendale_towards_2015, hendrycks_deep_2019, scheirer_2013}, adversarial examples \cite{szegedy_intriguing_2014}, and misclassifications, an inevitable part of any model since no model has perfect performance.  The main contribution of this work is the identification that the \textit{collective set} of erroneous inputs that cause failure can be jointly identified and separated from correctly processed inputs. 

The timely, preemptive detection of these failures is essential for preventing unreliable AI action based on incorrect predictions. An automated technique that broadly identifies when a model is in error is critically important for safe AI \cite{amodei_concrete_2016}. This need is particularly relevant now, given the increasingly ubiquitous deployment of deep learning models in real-world applications such as autonomous vehicles and medical devices.



\begin{figure*}[hptb]
\centering
\includegraphics[width=.99\linewidth]{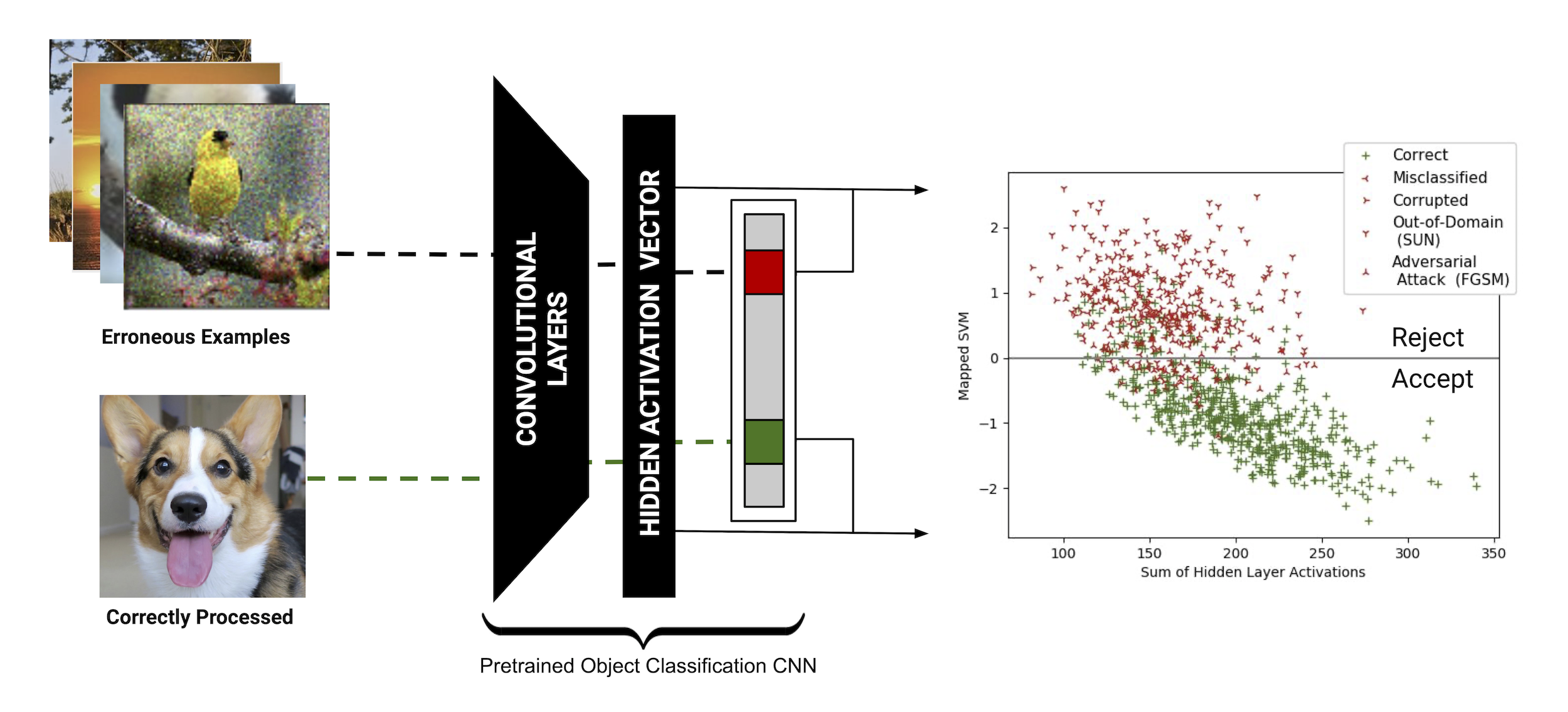}
\caption{Erroneous inputs cause neural network's to incorrectly classify images. However, these inputs can be preemptively identified with a simple linear SVM that is trained on the network's internal activations (i.e., features) that occur in response to the image. Specifically, the SVM is trained on features from the hidden activation vector and the softmax vector of the network. Using this technique, erroneous inputs are detected with an AUROC of 0.973 on CIFAR-10, 0.957 on Tiny ImageNet, and 0.941 on ImageNet. Above is the optimal hyperplane for CIFAR10 mapping erroneous examples against correct examples. }
\label{methods_teaser}
\end{figure*}

Given the breadth of erroneous input variations, it is unsurprising that each of these inputs has been traditionally tackled as a separate, individual problem requiring a unique, specialized solution. For example, out-of-distribution detection identifies when images are outside of a model's training (i.e., unseen classes)~\cite{devries_learning_2018, hendrycks_deep_2019, lee_training_2018, liang_enhancing_2018, scheirer_2013} (additional related work on erroneous input detection is discussed in Section \ref{related_work}). 
In our study of these seemingly disparate phenomena we noticed a common theme: the network's internal activation patterns were notably distinct when stimuli were \textit{correctly} processed. We hypothesized that erroneous inputs could be collectively separated from correctly processed inputs with a linear SVM trained on these features.

We test our hypothesis with three pre-trained image classification models, each pre-trained on a separate dataset, and find erroneous inputs can be broadly detected as a collective group and generalize to a multitude of alternative datasets. Further, we find that our detection technique is robust to four types of adversarial attacks. To our knowledge, this work is the first to consider the feasibility of broadly detecting four distinct types of erroneous inputs. Specifically, we find that the model's internal activations in the softmax layer and final hidden layer are sufficient for automatically filtering incorrectly processed inputs, as shown in Figure \ref{methods_teaser}. 

In summary, we make the following contributions:
\begin{itemize}
\item This is the first research showing that the four distinct types of erroneous inputs (see Figure \ref{teaser}), which neural networks will fail to correctly classify, can be \textit{jointly} separated from correctly classified images using the network's activations in response to the input. 
\item We achieve results comparable to  state-of-the-art techniques by considering the activations of both the softmax and final hidden layer of common pre-trained neural networks.  

\item We show how this method can be used as an auxiliary technique in existing anomaly detection models to enhance performance.
\end{itemize}

\section{Related Work} \label{related_work}

A variety of works have studied automatically detecting inputs that will cause neural networks to fail. In this section, we discuss each of the four subfields that compose erroneous inputs, and previously established connections between inputs. 

\textbf{Image Corruption}
Standard neural networks perform poorly on corrupted examples. Image corruption is an unintentional side-effect of any visual system: corruptions range from image quality degradation to lighting changes. In particular, networks have shown to be susceptible to Gaussian noise, blur, pixelation, and JPEG compression~\cite{dodge_study_2017,hendrycks_benchmarking_2019}. In real-world scenarios, like autonomous driving, models need to be robust to corruption from weather and debris~\cite{chaabane2019looking,michaelis_benchmarking_2019}; some research even proposed that networks should be explicitly evaluated on their robustness to corruption~\cite{RichardWebsterPsyPhy2019}. 

\textbf{Out-of-Distribution Inputs (OoD)}
\label{ood}
An OoD example occurs when a model is presented with data outside of its training paradigm. In this work, we defined OoD examples as originating from datasets other than the dataset the baseline model was trained on, although we controlled for class ambiguity (for example, if a dataset was trained on dogs, we controlled for wolves in OoD data). 

Hendrycks and Dietterich~\cite{hendrycks_baseline_2018} established a baseline for OoD detection by thresholding the Maximum Softmax Probability (MSP). Results were presented on several datasets in different neural network modals such as computer vision and natural language processing.
Other works trained neural networks to reject out-of-distribution data with auxiliary branches~\cite{devries_learning_2018}, auxiliary datasets~\cite{hendrycks_deep_2019}, GAN examples~\cite{lee_training_2018}, and other train-time techniques~\cite{bevandic2018discriminative,geifman_selectivenet_2019}. A notable cross-domain technique was proposed by Liang \etal~\cite{liang_enhancing_2018}, who used adversarial perturbations on input data to discriminate in-distribution from OoD examples. 

Further, Bendale and Boult proposed OpenMax~\cite{bendale_towards_2015} , a network layer that use the penultimate activation vector of neural networks and propose OpenMax, a network layer to estimate the probability of being from an unknown class using Meta Recognition. 
Specifically, for a given example, OpenMax takes the activation vector mean from each correctly predicted class to compute a Weibull distribution to determine the probability that a new example is OoD. They successfully detect out-of-distribution images and some types of adversarial attacks. Their out-of-distribution set contains fooling images labeled with high-confidence in standard softmax models, similar to our ImageNet-O set referenced section \ref{ood_datasets}.  Additional work by Rozsa et. al. \cite{rozsa_adversarial_2017} proposed an adversarial attack against OpenMax, which we describe in Section \ref{adv}.


\textbf{Adversarial Attacks}
\label{adv}
Adversarial examples occur when small, meaningful changes to clean data alter a network's prediction. Szegedy \etal\cite{szegedy_intriguing_2014} highlighted these limitations, giving rise to the field of adversarial attacks. Since then, numerous methods have been proposed to generate adversarial attacks~\cite{carlini_towards_2017, goodfellow_explaining_2015, moosavi-dezfooli_deepfool_2016,papernot_limitations_2015}. Several works have attempted to defend against adversaries~\cite{lu_safetynet_2017,metzen_detecting_2017}, however, the field largely conforms to a classic ``white hat"/``black hat" paradigm (i.e., solutions are often temporary because new attacks are created to overcome the defensive solution). 

Researchers have identified the upper bounds of adversarial robustness techniques for certain distributions~\cite{fawzi_adversarial_2018,gilmer_adversarial_2018,mahloujifar_curse_2018}. In particular, Fawzi \etal~\cite{fawzi_adversarial_2018} suggested a strong relation between adversarial robustness and the linearity of a classifier in latent space. Gilmer~\etal~\cite{gilmer_adversarial_2018} showed that when a model misclassifies even a fraction of inputs, it can be exposed to adversarial perturbations of size $O(1/\sqrt{n}$), where $n$ is the number of input dimensions. 

Our work finds that adversarial inputs, generated using a variety of popular methods, manifest linearly separable activation's we can use to detect when models will fail. While we test our model on a variety of adversarial methods, attacks on the penultimate layers features will likely cause our model to fail.  One example of this is LOTS \cite{rozsa_adversarial_2017}, an algorithm designed specifically to break OoD method OpenMax.   The algorithm modifies an image in order to mimick the features of the OpenMax.

\textbf{Misclassified Examples and Thresholding}
Misclassified examples occur when a model fails to predict an instance of a class, despite the model being trained to predict that class. Thresholding is a common technique to detect misclassified data, and has a rich history of research \cite{chow1970, chow_optimum_1957}, however, we believe we are the first to consider multiple layers of the network for thresholding.

Hendrycks and Dietterich~\cite{hendrycks_baseline_2018} provide thresholding results for misclassified examples on contemporary datasets, while \cite{geifman_selective_2017} presented results of the softmax response from a risk perspective. In recent years, thresholding techniques have been applied to detect adversarial examples~\cite{lu_safetynet_2017, pang_towards_2018} and OoD examples~\cite{hendrycks_baseline_2018}. Lu \etal~\cite{lu_safetynet_2017} provide the approach most similar to our own, utilizing layers toward the end of deep learning models to detect adversarial examples with SVM-RBF classifiers. Our work also shows that some adversarial examples can be identified using these features; however, we achieve higher accuracy by considering the hidden layers as well as the softmax layer.

\textbf{Connecting Erroneous Inputs}
\label{connecting}
Recent research has found similarities in the above areas. Hendrycks and Gimpel~\cite{hendrycks_baseline_2018} showed that both misclassified and out-of-distribution examples can be detected by utilizing the maximum softmax probability. Ford \etal~\cite{ford_adversarial_2019} proposed a similarity between the fields of adversarial and corrupted examples, showing empirical and theoretical evidence that these two fields are manifestations of the same phenomena. Another notable cross-domain technique was proposed by Liang \etal~\cite{liang_enhancing_2018}, who used adversarial perturbations on input data to discriminate in-distribution from OoD examples. Finally, Rozsa and Boult~\cite{rozsa_improved_2019} argued that adversarial perturbations exist in open space, contrary to popular belief that they exist near training samples. 


The above research highlights the distinct work being done in each of these four fields, and previous work that has identified connections between some of these fields. To our knowledge, our work is the first to fully explore the link between all of these areas, and the first to detect them with a single approach.

\section{Method}

We first formally define the domain of erroneous inputs (Section \ref{domain}) and set up the experimental design (Section \ref{experimental_setup}), enumerating the pre-trained models and datasets used in experiments. Finally, we define the methods for training and testing (Section \ref{binary_classification_details}), evaluation metrics (Section \ref{testing_and_evaluation_metrics}), and other details (Section \ref{other}) such as dataset sizes. 

\subsection{Defining the Domain} \label{domain}
We defined our domain in the context of detecting when input data to a visual classification model, $f(\mathcal{X}) \rightarrow \mathcal{Y} $, was classified incorrectly. Incorrect classifications in our experiments encompassed out-of-distribution classes $(\mathcal{D}_{out})$, adversarial examples $(\mathcal{A})$, corrupted examples ($\mathcal{C}$), and misclassified in-distribution data ($\mathcal{M}$). $\mathcal{M}$ could be considered any input from in-domain dataset $\mathcal{D}_{in}$ where output $\mathcal{Y}_{predicted}\neq \mathcal{Y}_{actual}$. 

In our experiments, we train a binary classifier, $h(\mathcal{X}')$, to detect erroneous examples \{$\mathcal{D}_{out}$, $\mathcal{A}$, $\mathcal{C}$, $\mathcal{M}$\}, where $\mathcal{X}'$ is the activations of the penultimate and final layers of pretrained model $f$ for image $\mathcal{X}$. In $h(\mathcal{X}')$, we consider $\mathcal{D}_{out}$, $\mathcal{A}$, $\mathcal{C}$, and $\mathcal{M}$ as belonging to the positive class, which were in turn classified against the correctly predicted in-domain dataset $\mathcal{D}_{in}$, where $\mathcal{Y}_{predicted}= \mathcal{Y}_{actual}$.

To build the training and tests set for our binary classifier we use validation and test sets from the dataset under test.  Training data is completely disjoint from our test data.  Details in Section \ref{experiment_setup}.  
Our training set was built such that $f(\mathcal{A}) \neq \mathcal{Y}_{actual}$ and $f(\mathcal{C}) \neq \mathcal{Y}_{actual}$ --- we only used samples which the pretrained neural network classified incorrectly. 
When an adversarial or corrupted example is classified correctly, we do not want to remove the example from the baseline models purview. 
Further, adversarial examples were generated from inputs the model originally predicted correctly, i.e. $f(x)=\mathcal{Y}_{actual}$ and $f(x \rightarrow \mathcal{A}) \neq \mathcal{Y}_{actual}$. Finally, it is implied that $f(\mathcal{D}_{out})$ and $f(\mathcal{M})$ would produce invalid results, so each example will be taken from these datasets.

It should be noted that we did not classify in-domain versus out-of-domain data because we consider the set of everything except $\mathcal{D}_{out}$ to be in-domain: \{ $\mathcal{D}_{in}$, $\mathcal{A}$, $\mathcal{C}$, $\mathcal{M}$\} $\in $ $\mathcal{D}$ while $\mathcal{D}_{out}$ $\notin $ $\mathcal{D}$. Further, our initial experiments were set up to classify correct example sets from each of the erroneous examples sets, i.e. classifying correct examples against the set of misclassified examples. Further, we combine the erroneous example sets, and jointly classified the data from correct examples.


\subsection{Experimental Setup} \label{experimental_setup}
For our experiments we used several common baseline datasets and models for image classification. The baseline models were fed various correct data and erroneous data and the activations for both the softmax output and the penultimate layer of the model were collected for use in our detection models. The baseline models were not altered in any way during our experiments. 

\subsubsection{Baseline Datasets and Models}

\label{baseline_models}
Below are the summaries of the pre-trained models used for each dataset under test. 
For uniformity, all images passed to the pre-trained models were resized and normalized the same way. 

\textbf{CIFAR-10} contains 32 x 32 colored images of 10 different classes of objects~\cite{Krizhevsky_Hinter_2009}. The dataset has 50,000 training images and 10,000 testing images. Our CIFAR-10 model was trained using the ResNet50 architecture and open-sourced by~\cite{madry_towards_2019}.

\textbf{Tiny ImageNet} is a 200-class subset of the ImageNet dataset where images were cropped and resized to a resolution of 64 x 64. Bounding box information was used in the image cropping~\cite{tiny_imagenet}. Our Tiny ImageNet model is a pre-trained WideResNet~\cite{zagoruyko_wide_2017}. The trained model was open-sourced by~\cite{hendrycks_benchmarking_2019}.

\textbf{ImageNet} consists of 1000 classes of objects~\cite{ILSVRC15}, with varied dimensions and resolutions. Our work used the validation dataset to produce examples for our detection model. We utilized the default pre-trained ResNet50 model in the Pytorch library for our ImageNet experiments. 


\subsubsection{Out-of-Distribution Datasets}
\label{ood_datasets} 
We use several commonly used OoD datasets:    

    \textbf{CIFAR-100} contains 32 x 32 colored images of 100 different classes of objects~\cite{Krizhevsky_Hinter_2009}. We filtered out classes similar to those found in CIFAR-10 to ensure all classes were truly OoD, leaving 74 classes. Details on excluded classes are in the appendix. 

\textbf{SVHN} We use the test set from the Street View House Numbers dataset \cite{netzer_reading_nodate}, which contains colored numbers with class labels 0 to 9. 

\textbf{The Scene UNderstanding dataset (SUN)} contains images of scenes with varying resolutions~\cite{xiao_sun_2010}. 

\textbf{Places365} contains 365 classes of scenes with varying image resolutions. We use the high-resolution validation set for use in our Tiny ImageNet and ImageNet models~\cite{zhou_places_2018}. 

\textbf{ImageNet-O} dataset was constructed from sampled images from ImageNet-22K~\cite{hendrycks_natural_2020}. First, images overlapping with classes in ImageNet-1K were filtered out. Next, they retained images that a ResNet50 ImageNet-1K model classified with high softmax probability. Finally, they hand-selected a subset of high-quality images.

\subsubsection{Corruption Datasets}
\textbf{CIFAR-10-C, Tiny ImageNet-C, ImageNet-C} are image corruption datasets for the purpose of testing model robustness~\cite{hendrycks_benchmarking_2019}. Each corrupted dataset included 15 common visual corruptions such as Gaussian noise, blur, and digital distortions. Each type of corruption had five levels of 'severity' for a total of 75 distinct corruptions per image. We used each corruption and severity type in our models. 


\subsubsection{Adversarial Attack Methods} \label{adversarial inputs}

 We test our method on four different adversarial attacks, however, we do not claim this method to be robust to \textit{any} attack or modified hyperparameters of the attacks used.  Section \ref{high_confidence_aa} looks at this further. Further, we looked at the first 20 generated examples of each adversarial dataset to verify the perturbed images were realistic.

White-box attack methods are used to generate adversarial examples. White-box methods assume the attacker has full access to the learning model.  Next we describe each attack used in our experiments.

\textbf{Fast Gradient Sign Method (FGSM)} was an attack introduced by Goodfellow et al.~\cite{goodfellow_explaining_2015}. It adjusts the inputs to maximize the loss based on the back-propagated gradients of the predicted value: 
$f(\mathcal{A})=\mathcal{X}+\epsilon sign( \nabla_x J(\theta, \mathcal{X}, \mathcal{Y}))$. FGSM examples were generated using $\epsilon$=0.01. 
 
\textbf{Carlini and Wagner} \textbf{$\textbf{l}_\textbf{2}$} \textbf{Attack (C \& W $l_{2}$)} 
searches for low distortion in the $l_2$ metric \cite{carlini_towards_2017}. Notable in the Carlini and Wagner attacks is $\kappa$, which allows a confidence level for the adversarial example to be calibrated. This means examples can be generated that the neural network predicts incorrectly with high probability. Examples in our main experiments were generated with $\kappa=0$ --- we did not search for a perturbed input which satisfied a specific confidence. 
 
 \textbf{Carlini and Wagner} \textbf{$\textbf{l}_\textbf{$\infty$}$} \textbf{Attack (C \& W $l_{\infty}$)} was a modified version of the C\&W $l_2$ attack. It controls the $l_{\infty}$ norm, i.e. the maximum perturbation applied to any pixel \cite{carlini_towards_2017}. 
 
\textbf{Projected Gradient Descent (PGD)} finds the perturbation that maximizes the loss of a model on an input, and, after each iteration, it projects the perturbation onto an ${l}_p$ ball of radius $\epsilon$ while also clipping values so they fall within a permitted range~\cite{madry_towards_2019} .

Carlini and Wagner $l_{2}$, $l_{\infty}$ attacks and PGD attack examples were generated using the adversarial-robustness-toolbox with default parameters~\cite{art2018}. We used the PyTorch implementation of the FGSM attack. 

\textbf{ImageNet-A} is a dataset of 'natural adversaries'. ImageNet-A was created by taking examples from the ImageNet dataset and removing examples the model predicted correctly. Then, a subset of high-quality images were hand-selected~\cite{hendrycks_natural_2020}. We included this dataset as an extension to our misclassified experiments.

\begin{table*}[t]

\begin{center}
\caption{CIFAR-10 and Tiny ImageNet experiments for MSP, Outlier Exposure, Outlier Exposure + Linear SVM (Ours), and Linear SVM (Ours).  Results are presented using five-fold cross validation for each model.  All error rates are near zero (less than 0.01).  Best results for each row are bolded. }
\vskip 0.1in
\small
\setlength{\tabcolsep}{6pt}
\renewcommand{\arraystretch}{1.1} 
\label{c10_tinyimagenet}
\begin{tabular}{lccccccccccr}

\toprule

 \textbf{Test Type} & & \multicolumn{2}{c}{MSP \cite{hendrycks_baseline_2018}} & \multicolumn{2}{c}{OE \cite{hendrycks_deep_2019}} &\multicolumn{2}{c}{OE+Linear SVM (Ours)} & \multicolumn{2}{c}{Linear SVM (Ours)}\\
 \midrule
\textbf{CIFAR-10} & \textbf{Detail} & AUROC &AUPR& AUROC &AUPR& AUROC &AUPR& AUROC &AUPR \\
 Misclassified & - & .930&.905&.927&.908&.935&.910&\textbf{.939}&\textbf{.911}\\
 Out-of-Distribution & Sun & .928&.912&.998&.998&\textbf{1.0}&\textbf{1.0}&.989&.989\\
 Out-of-Distribution & CIFAR100 & .925&.908&.992&.991&\textbf{1.0}&\textbf{1.0}&.970&.966\\
  Out-of-Distribution & SVHN & .954&.937&.998&.996&\textbf{1.0}&\textbf{1.0}&\textbf{1.0}&\textbf{1.0}\\
   Corrupted & CIFAR10-C & .945&.926&.961&.959&.975&.975&\textbf{.982}&\textbf{.982}\\
    Adversarial & FGSM & .830&.825&.998&.998&\textbf{1.0}&\textbf{1.0}&.980&.980\\
     Adversarial &  C \& W ${l}_{2}$ & .982&.973&.752&.747&.891&.891 &\textbf{
     .991}&\textbf{.989}\\
  Adversarial &  C \& W ${l}_{\infty}$ & .916&.893&.817&.808&.910&.906&\textbf{1.0}&\textbf{
  1.0}\\
    Adversarial & PGD & .999&.999&.989&.996&\textbf{1.0}&\textbf{1.0}&\textbf{1.0}&\textbf{1.0}\\
    \textbf{Combined} & &.925&.915&.871&.908&\textbf{.973}&\textbf{.977}&.969&.970 \\
 \toprule
 \label{ hidden+sorted}
 \textbf{Tiny ImageNet}\\
 Misclassified & - &\textbf{.860}&\textbf{.834}& .846&.809&.820&.803&.847&.822\\
 Out-of-Distribution & Sun & .876&.864&\textbf{.999}&\textbf{.999}&.999&.998&.994&.993\\
  Out-of-Distribution & Places365 & .882&.873&.993& \textbf{.999}&\textbf{.999}&.998&.992&.991 \\
   Corrupted & TinyImageNet-C &.895&.885&.996&.996&\textbf{.998}&\textbf{.997}&.994&.993 \\
    Adversarial & FGSM & .998&.999&.720&.699&.984&.983&\textbf{.999}&\textbf{.999}\\
     Adversarial &  C \& W ${l}_{2}$ & .908&.835&.823&.794&\textbf{.994}&\textbf{.994}&.967&.960\\
  Adversarial &  C \& W ${l}_{\infty}$ & .855&.782&.815&.785&\textbf{.995}&\textbf{.995}&.887&.858\\
    Adversarial & PGD & .990&.991&.743&.718&.996&.998&\textbf{1.0}&\textbf{1.0} \\
\textbf{Combined} & & .886&.870&.828&.863&\textbf{.957}&\textbf{.964}&.931&.933\\
 \bottomrule
\end{tabular}

\end{center}
\end{table*}

\subsection{SVM Model} \label{binary_classification_details}
We employed a linear Support Vector Machine (SVM) classifier for our experiments, where examples (\{${{x_1}}$\}, $y_1$), ..., (\{${{x_n}}$\}, $y_n$) were trained to maximize the hyperplane between the groups y=0 and y=1.  Correct examples were the base class (y=0). 

While MSP and OE use a model-less ROC during evaluation, we evaluate our detection model using five-fold cross validation.  We found that linear SVM's were able to generalize better in high dimensions than other SVM kernels.   


To generate features \{${{x_1}}$\}...\{${{x_n}}$\} for our SVM, we take values from the softmax output of the baseline model, sort them, and concatenate them onto the fully-connected activations from the penultimate layer. For ResNet50 pretrained models (CIFAR10 and ImageNet), the penultimate layer is 2048 values, while Tiny ImageNet has a penultimate layer of size 128.

We present results in this paper with a balanced dataset: each model contains the same number of correct and erroneous examples.

\begin{algorithm}
\caption{SVM Training for our hidden activation + sorted softmax algorithm}
\begin{algorithmic}[1]
\REQUIRE Pretrained object classification model, $f(\mathcal{X})$, where $f_p(\mathcal{X})$=penultimate layer of $f(\mathcal{X})$ and $f_{ss}(\mathcal{X})$=sorted softmax output of $f(\mathcal{X})$, i.e. ($f(\mathcal{X})$). 
\REQUIRE Known erroneous example set\{ $\mathcal{D}_{in}$, $\mathcal{A}$, $\mathcal{C}$, $\mathcal{M}$\} and known correct example set \{$\mathcal{C}$\}. Erroneous examples have y-label 1 in SVM model, while correct examples have y-label 0 in SVM. 
\ENSURE Size of correct example set is equal to size of erroneous example set: $\lvert\{ \mathcal{C}\} \rvert= \lvert\{ \mathcal{D}_{in}, \mathcal{A}, \mathcal{C}, \mathcal{M}\}\rvert$

\FOR{\textbf{i} in \{$\mathcal{C}, \mathcal{D}_{in}$, $\mathcal{A}$, $\mathcal{C}$, $\mathcal{M}$\}}
\STATE $features \leftarrow [p(i), ss(i)]$
\ENDFOR
\STATE SVM Fit $\{features_i,...features_n\}$
\RETURN SVM detection model
\end{algorithmic}
\end{algorithm}

\subsection{Testing and Evaluation Metrics} \label{testing_and_evaluation_metrics}
 We evaluate binary detection tasks using three metrics: area under the receiver operating characteristic curve (AUROC), area under the precision-recall curve (AUPR), and false positive rate at N\% true positive rate (FPR$\mathcal{N}$).

AUROC plots the true positive rate (TPR) against the false positive rate (FPR). Random classifiers score 50\% while perfect classifiers achieve 100\%. 
AUPR is another cumulative distribution function which plots the precision (True Positive)/(True Positive + False Positive) versus the recall (True Positive)/(True Positive + False Negative). 

For our ImageNet model we also present FPR$\mathcal{N}$ scores, also used by~\cite{hendrycks_deep_2019, liang_enhancing_2018, liu_open_2018}. The FPR$\mathcal{N}$ calculates the false positive rate at a set True Positive Rate. 
We use TPR of 95\%, similar to past papers. 

\subsection{Other Experiment Details}
\label{other}

\textbf{Dataset sizes}  Sizes generally ranged from 3,500 examples up to 50,000 examples.  
There were two exceptions to this rule: our CIFAR10 pretrained model only misclassified 475 images on the test set, and TinyImageNet FGSM attacks rendered only 1,764 bad examples.  Details in the appendix. 

\textbf{Baseline detection models} We test our erroneous datasets on two baselines: MSP \cite{hendrycks_baseline_2018} and Outlier Exposure \cite{hendrycks_deep_2019}.  For MSP, we use the pretrained models as described in Section \ref{baseline_models}.  We use the same pretrained models for our main experiments, the columns labeled ``Linear SVM" in Tables \ref{c10_tinyimagenet} and \ref{imagenet}.  As an additional measure, we test the datasets on Outlier Exposure by using pretrained models open sourced by the researchers.  The models were trained with a modified loss function which encourages out-of-distribution examples to fit a uniform distribution.  An important note is that they expose the pretrained model to OoD examples.  Further, the choice of OoD exposure was important, for instance, corrupting in-distribution examples with noise did not perform well.  As a result, the models were trained with realistic OoD data from 80 Million Tiny Images dataset and ImageNet22k.  

\label{experiment_setup}

\section{Results}

We present our findings: First, Section \ref{single} breaks down the results by pretrained models and datasets, then, Section \ref{combined_model} presents the combined results across erroneous inputs.

\subsection{Single Erroneous Dataset Results}
\label{single}

In Table \ref{c10_tinyimagenet} we compared Maximum Softmax Probability (MSP) \cite{hendrycks_baseline_2018} and Outlier Exposure (OE) \cite{hendrycks_deep_2019} against our OE model and our linear SVM, focusing on two pretrained models: CIFAR10 and Tiny ImageNet. All results are validated with 5-fold cross validation.  Results show that our methods achieve state-of-the-art results on all experiments except for misclassified inputs on the Tiny ImageNet model. By applying our method of using the hidden activations as well as the sorted softmax to linearly discriminate correct examples from erroneous examples, we are able to outperform OE and MSP on 15 of 17 datasets.  


In Table \ref{imagenet}, we present our results for ImageNet examples in a pretrained ResNet model. Our linear SVM was trained and evaluated with five-fold cross validation. In this case, we only evaluated our method against the baseline AUROC curve of the MSP.  Across all of our tests, our models easily outperformed MSP.

\begin{table*}[t]
\begin{center}

\caption{ImageNet pretrained model experiments for MSP as well as our Linear SVM model. Up arrows indicate when a higher score is better, while down arrows indicate when a lower score is better. OoD stands for out-of-distribution, Cor. stands for corrupted, and Adv. stands for adversarial attack. See appendix for all experimental results. }
\vskip 0.1in
\small
\setlength{\tabcolsep}{6pt}
\renewcommand{\arraystretch}{1.1} 
\label{imagenet}
\begin{tabular}{lccccccccr}

\toprule
 \textbf{ImageNet} && \multicolumn{3}{c}{MSP \cite{hendrycks_baseline_2018}} &\multicolumn{3}{c}{Linear SVM (Ours)} \\
 \midrule
\textbf{Test Type} & \textbf{Detail} & AUROC $\uparrow$ &AUPR $\uparrow$& FPR (95\%) $\downarrow$  &AUPR $\uparrow$& AUROC $\uparrow$ &FPR (95\%) $\downarrow$ \\
 Misclassified &-& .853&.833&.506&\textbf{.898}&\textbf{.868}&\textbf{.395}\\
  Misclassified & ImageNet-A & .908&.910&.423&\textbf{.940}&\textbf{.940}&\textbf{.052}\\
 Out-of-Distribution & Sun & .830&.822&.640&\textbf{.983}&\textbf{.981}&\textbf{.021} \\
 Out-of-Distribution & ImageNet-O &.656&.552&.617&\textbf{.897}&\textbf{.873}&\textbf{.035}\\
 Out-of-Distribution & Places365 &.846&.843&.618&\textbf{.983}&\textbf{.981}&\textbf{.015}\\
 Corrupted & ImageNet-C &.927&.924&.326&\textbf{1.0}&\textbf{1.0}&\textbf{0.0}\\
  Adversarial & FGSM &.949&.944&.222&\textbf{1.0}&\textbf{1.0}&\textbf{0.0}\\
   Adversarial & C \& W ${l}_{\infty}$& .870&.802&\textbf{.320}&\textbf{.875}&\textbf{.847}&.444\\
   Adversarial & C \& W ${l}_{2}$ &.890&.812&\textbf{.223}&\textbf{.930}&\textbf{.911}&.236\\
     Adversarial & PGD& .990&.990&.030&\textbf{.996}&\textbf{.997}&\textbf{.001}\\
\textbf{Combined} & &.881&.876& .472&\textbf{.941}&\textbf{.947}&\textbf{.277} \\
 \bottomrule
\end{tabular}

\end{center}
\end{table*}

\subsection{Joint Erroneous Example Detection} \label{combined_model}

Section \ref{single} showed how our linear SVM accurately filtered each individual erroneous input type from correctly processed inputs. In this section, we investigate the combined condition ('Combined' row in Tables \ref{c10_tinyimagenet} and \ref{imagenet}), where the full set of erroneous inputs are detected by a single linear model. Here, our experiments unequivocally show that, across all of our models, the full set of erroneous inputs could be detected with comparable performance to models trained to detect individual types of erroneous inputs (AUROCs between $0.941$ and $0.973$).

A manual analysis of model failures found we successfully identified erroneous inputs except examples from PGD attacks. For PGD, detection failed because inputs tended to have higher maximum softmax probabilities (PGD inputs had nearly $100\%$ probability, on average) than the softmax probabilities of correctly classified examples ($80\%$ MSP average for ImageNet, $86\%$ MSP average for Tiny ImageNet). Inversely, all other erroneous example sets had lower softmax probabilities (between $25\%$ and $93\%$ average). Users of this technique will need to be mindful of cases like this, since PGD examples will need their own model. However, such a model can easily separate correct examples, as shown in the PGD experiments in Tables \ref{c10_tinyimagenet} and \ref{imagenet}.




\section{Further Investigation}
\label{blessing}
We examined four additional scenarios in the following section: linear separation among erroneous example sets,  high-softmax erroneous examples, left-out erroneous datasets, and finally corrupted examples with correct predictions.  Each of the experiments show evidence of the beneficial properties of high-dimensional data, otherwise known as the \textit{'Blessing of Dimensionality'} \cite{kainen_utilizing_1997}.  See discussion in Section \ref{discussion} for more details.

\textbf{Discriminability of Erroneous Sets} 
We investigated whether erroneous datasets occupied the same feature space by training an SVM to linearly separate erroneous inputs. Our results, shown in Table \ref{erroneous_seperability}, indicate that each erroneous set generally occupies a unique feature space in \textit{
high dimensions}. However, in lower dimensions (i.e., the single dimension MSP), the results were close to random. This underscores the importance of considering the latent vector in our experiments, and why previous works have only found links between particular pairs of erroneous input types, while we are able to jointly separate all erroneous inputs from correctly processed inputs. 

\label{high_confidence_aa}
\textbf{High Softmax Erroneous Examples} 
We explored if our method was susceptible to high softmax erroneous examples. Specifically, we filter our CIFAR10 datasets to retrieve examples where the maximum softmax probability is greater than 0.99999.  
Our filtering yields 342 erroneous examples from each type (we exclude PGD, as discussed in Section \ref{combined_model}). 
MSP has an AUROC of .513 and an AUPR of .532, or nearly random. Our method achieved an AUROC of .922 ($\pm.026$) and an AUPR of .935 ($\pm.014$). Results positively indicate that the high-dimensional feature data performs substantially better than its single dimensional counterpart (MSP).

\begin{table}[h]
\caption{We test the separability of erroneous inputs on our CIFAR10 model with Maximum Softmax Probability and a Linear SVM.  The linear SVM model is trained on the penultimate layer as well as the sorted softmax output, and contains 2058 features.  Results indicate that each type of erroneous input manifests largely in it's own feature space in high dimensions, but is not separable in lower dimensions. Our experiments are performed with 5-fold cross validation, with error rates all below 0.015.   }
\vskip 0.1in
\begin{center}
\begin{small}
\begin{sc}
\label{erroneous_seperability}
\setlength{\tabcolsep}{1pt}
\renewcommand{\arraystretch}{1.1} 
\begin{tabular}{lccccccr}
\toprule
&&\multicolumn{2}{c}{MSP}&\multicolumn{2}{c}{Linear SVM}\\
Base&Second & AUROC & AUPR&AUROC&AUPR  \\
\toprule
Incorrect&FGSM&.634&.660&1.0&1.0 \\
Corr.&Incorrect&.530&.530&1.0&1.0 \\
FGSM&Corr.&.681&.641& .981&.983\\
Corr.&C-100&.538&.426&.935&.928 \\
C-100&Incorrect&.522&.500&1.0&1.0\\
FGSM&C-100&.634&.612&.988&.989 \\
\end{tabular}
\end{sc}
\end{small}
\end{center}
\vskip -0.1in
\end{table}

 \label{leaveoneout}
\textbf{Leave One Erroneous Set Out}
Our experiments show we can successfully filter erroneous inputs from correctly processed ones, but there may be additional kinds of erroneous inputs that we have not tested here. To investigate how robust our method would be against a potential ``unknown" erroneous input, we evaluated our method in a leave-one-out context, where we train a linear SVM on each erroneous set minus a single left out erroneous input type, and then test the model on the left out type. A model capable of discriminating new erroneous example types has substantial benefits in a real-world use case.

In our experiments, we use one dataset from each erroneous group from our CIFAR10 model: the misclassification dataset, FGSM attacks, SVHN, and CIFAR10-C. 

Our results indicate that our method will be able to generalize to new types of erroneous inputs, but at the cost of accuracy: most notably, unseen FGSM detection dropped to an AUROC of .822 and AUPR of .817, a substantial drop from .999 AUROC when FGSM inputs were part of training. Similiarly, unseen SVHN dropped slightly from an AUROC of 1.0 to .987 (AUPR = .988), Unseen corrupted were detected with .952 AUROC .958, compared with .998 AUROC originally. Interestingly, misclassified examples were detected at similar rate: .863 AUROC and  .855 AUPR.  

\textbf{Corrupted Example}
Finally, we experimentally evaluated how effectively we could separate correctly processed corrupt examples ($f(\mathcal{C}) = \mathcal{Y}_{actual}$) from erroneous examples. For this experiment, we used five-fold cross validation on our CIFAR-10 dataset. Experiments elicited strong results, with AUROC's between .956 and 1 and AUPR's from 0.955 to 1.  Interestingly, the dataset that performed the worst was incorrect corrupted images, with AUROC 0.956, indicating the model had most trouble discriminating similarly corrupted images.  MSP, on the other hand, has an AUROC of 0.813 and 0.794 when discriminating corrupted examples with correct predictions from other erroneous predictions.  

\section{Real-World Use Case}
To promote potential use of this model, we describe a simple production scenario for this algorithm:

An autonomous vehicle relies on a legacy machine learning algorithm $f(x)$ to determine its course of action, where $x$ is the video input from the vehicle's camera.  Specifically, $f(x)$ outputs a vehicle's next course of action (e.g. turning the wheel, stopping, etc.) based on visual cues from the road ahead.  The model was originally trained to be robust against erroneous inputs by a method such as OE \cite{hendrycks_deep_2019}, however, the model still fails in certain scenarios, such as: 1)  Natural adversarial images that the model misclassifies, such as the dog behind a bush pictured in Figure \ref{teaser}. 2)  The model receives corrupted input data, such as camera distortion in poor weather. 3)  The model receives an input which it has never seen before, such as an exotic animal.  4) New adversarial attacks on the camera system meant to fool the model.  

During testing and use of the autonomous driving software, engineers have flagged various erroneous inputs where $f(x)$ fails.  These inputs are compiled into a dataset, and the engineers follow our methods to train an SVM to delineate bad examples from inputs where $f(x)$ performs well.  The resulting model can be quickly deployed to alert drivers when \textit{new} erroneous inputs are present and the driver needs to take control of the vehicle.  This quick corrective mechanism adds safety measures to an existing production software. 


\begin{figure}[hptb]
\centering
\includegraphics[width=.99\linewidth]{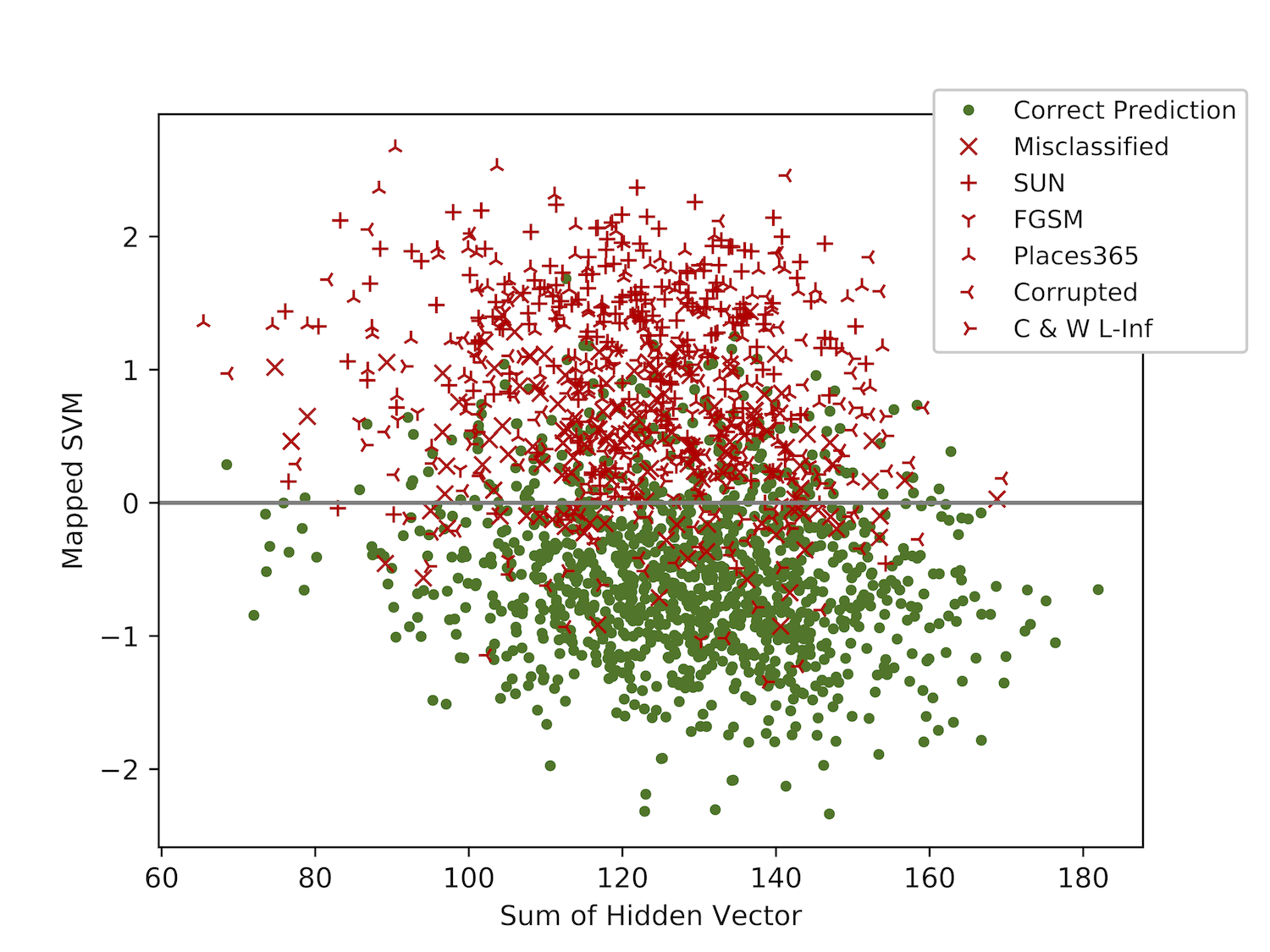} 

\caption{The hyperplane separating erroneous examples from correct examples in our Tiny ImageNet model. "0" on the y-axis represents the optimal hyperplane in a linear SVM classifier. 
To improve visualization, we only include a subset of the correct and erroneous example sets. }
\label{seperating_all_erroneous_inputs}
\end{figure}


\section{Discussion}
\label{discussion}

Erroneous inputs have been largely studied as distinct phenomena, however, our results show that these faulty inputs can be broadly detected by considering a model's internal behavior. 
We propose a new internal activation combination that allows for the broad detection  of faulty data including  corrupted, out-of-distribution, misclassified , and four types of adversarial attacks. 



We believe that the positive outcomes in this paper result from the beneficial properties of high-dimensional data. Contrary to the popular 'Curse of Dimensionality' \cite{bellman_dynamic_1958}, which argues that problems become more difficult in high-dimensions, Kainen coined the term 'Blessing of Dimensionality' describing scenarios in which complex data is more beneficial \cite{kainen_utilizing_1997}. 

Stochastic separation theorems recently introduced by Gorban and Tyukina \cite{gorban_stochastic_2017} formally established this phenomena, showing that in moderately high-dimensions we can achieve linear separability of sets with probability close to 1.  Further work by the researchers presented a similar experiment to our own \cite{gorban_highdimensional_2020} by using LDA to discriminate anomalies from correct data on a simple dataset.  These results are consistent with our experiments in the preceding sections.

A model capable of interpreting the breadth of all inputs is at best years away, and thus detection of `bad' data, which will cause models to fail, is essential for the safe use of any real-world system. We argue that by moving towards detection of bad data in the broader application of a learning based system, we can advance the reliability of machine learning models in real-world applications. 

\section*{Acknowledgement}
\noindent
This work was supported by the National Science Foundation award 2016714.

{\small
\bibliographystyle{ieee_fullname}
\bibliography{egbib}
}

\end{document}